\documentclass[10pt,twocolumn,letterpaper]{article}
\pdfoutput=1
\usepackage{wacv}
\usepackage{times}
\usepackage{epsfig}
\usepackage{graphicx}
\usepackage{amsmath}
\usepackage{amssymb}
\usepackage{multirow}
\usepackage{array}
\usepackage{arydshln}
\usepackage{paralist}

\usepackage{siunitx}
\usepackage{subcaption}
\usepackage{booktabs}
\usepackage{gensymb}
\usepackage[altpo]{backnaur}
\usepackage{appendix}

\newcolumntype{L}{>{$}l<{$}}
\newcolumntype{C}{>{$}c<{$}}
\newcolumntype{R}{>{$}r<{$}}

\newcommand{\nm}[1]{\textnormal{#1}}

\def\wacvPaperID{1066} 

\wacvfinalcopy 

\ifwacvfinal
\def\assignedStartPage{123} 
\fi
\ifwacvfinal
\usepackage[breaklinks=true,bookmarks=false]{hyperref}
\else
\usepackage[pagebackref=true,breaklinks=true,colorlinks,bookmarks=false]{hyperref}
\fi

\ifwacvfinal
\else
\fi

\begin{document}

\title{A Large-Scale, Time-Synchronized Visible and Thermal Face Dataset}

\author{Domenick Poster\renewcommand*{\thefootnote}{\arabic{footnote}}\footnotemark[1]$\;^,$\renewcommand*{\thefootnote}{\fnsymbol{footnote}}\footnotemark[1]
\and
\renewcommand*{\thefootnote}{\arabic{footnote}}
Matthew Thielke\footnotemark[2]
\and
\renewcommand*{\thefootnote}{\arabic{footnote}}
Robert Nguyen\footnotemark[2]$\;^,$\footnotemark[3]
\and
\renewcommand*{\thefootnote}{\arabic{footnote}}
Srinivasan Rajaraman\footnotemark[2]$\;^,$\footnotemark[3]
\and
\renewcommand*{\thefootnote}{\arabic{footnote}}
Xing Di\footnotemark[2]$\;^,$\footnotemark[4]
\and
\renewcommand*{\thefootnote}{\arabic{footnote}}
Cedric Nimpa Fondje\footnotemark[5]
\and
\renewcommand*{\thefootnote}{\arabic{footnote}}
Vishal M. Patel\footnotemark[4]
\and
\renewcommand*{\thefootnote}{\arabic{footnote}}
Nathaniel J. Short\footnotemark[2]$\;^,$\footnotemark[3]
\and
\renewcommand*{\thefootnote}{\arabic{footnote}}
Benjamin S. Riggan\footnotemark[5]
\and
\renewcommand*{\thefootnote}{\arabic{footnote}}
Nasser M. Nasrabadi\footnotemark[1]
\and
\renewcommand*{\thefootnote}{\arabic{footnote}}
Shuowen Hu\footnotemark[2]
\and
\renewcommand*{\thefootnote}{\arabic{footnote}}\footnotemark[1]~~West Virginia University, 395 Evansdale Dr., Morgantown, WV 26506\\
\renewcommand*{\thefootnote}{\arabic{footnote}}\footnotemark[2]~~DEVCOM Army Research Laboratory, 2800 Powder Mill Rd., Adelphi, MD 20783\\
\renewcommand*{\thefootnote}{\arabic{footnote}}\footnotemark[3]~~Booz Allen Hamilton, 8283 Grennsboro Dr., McLean, VA 22102\\
\renewcommand*{\thefootnote}{\arabic{footnote}}\footnotemark[4]~~Johns Hopkins University, 3400 N. Charles Street, Baltimore, MD 21218\\
\renewcommand*{\thefootnote}{\arabic{footnote}}\footnotemark[5]~~University of Nebraska-Lincoln, 1400 R St, Lincoln, NE 68588\\
\footnotemark[1]~~\emph{Corresponding authors: dposter@mix.wvu.edu}
}

\maketitle

\begin{abstract}
    
   Thermal face imagery, which captures the naturally emitted heat from the face, is limited in availability compared to face imagery in the visible spectrum. To help address this scarcity of thermal face imagery for research and algorithm development, we present the DEVCOM Army Research Laboratory Visible-Thermal Face Dataset (ARL-VTF). With over 500,000 images from 395 subjects, the ARL-VTF dataset represents, to the best of our knowledge, the largest collection of paired visible and thermal face images to date. The data was captured using a modern long wave infrared (LWIR) camera mounted alongside a stereo setup of three visible spectrum cameras. Variability in expressions, pose, and eyewear has been systematically recorded. The dataset has been curated with extensive annotations, metadata, and standardized protocols for evaluation. Furthermore, this paper presents extensive benchmark results and analysis on thermal face landmark detection and thermal-to-visible face verification by evaluating state-of-the-art models on the ARL-VTF dataset.
\end{abstract}

\section{Introduction}
\begin{figure}[tb]
    \centering
    \includegraphics[width=.95 \linewidth]{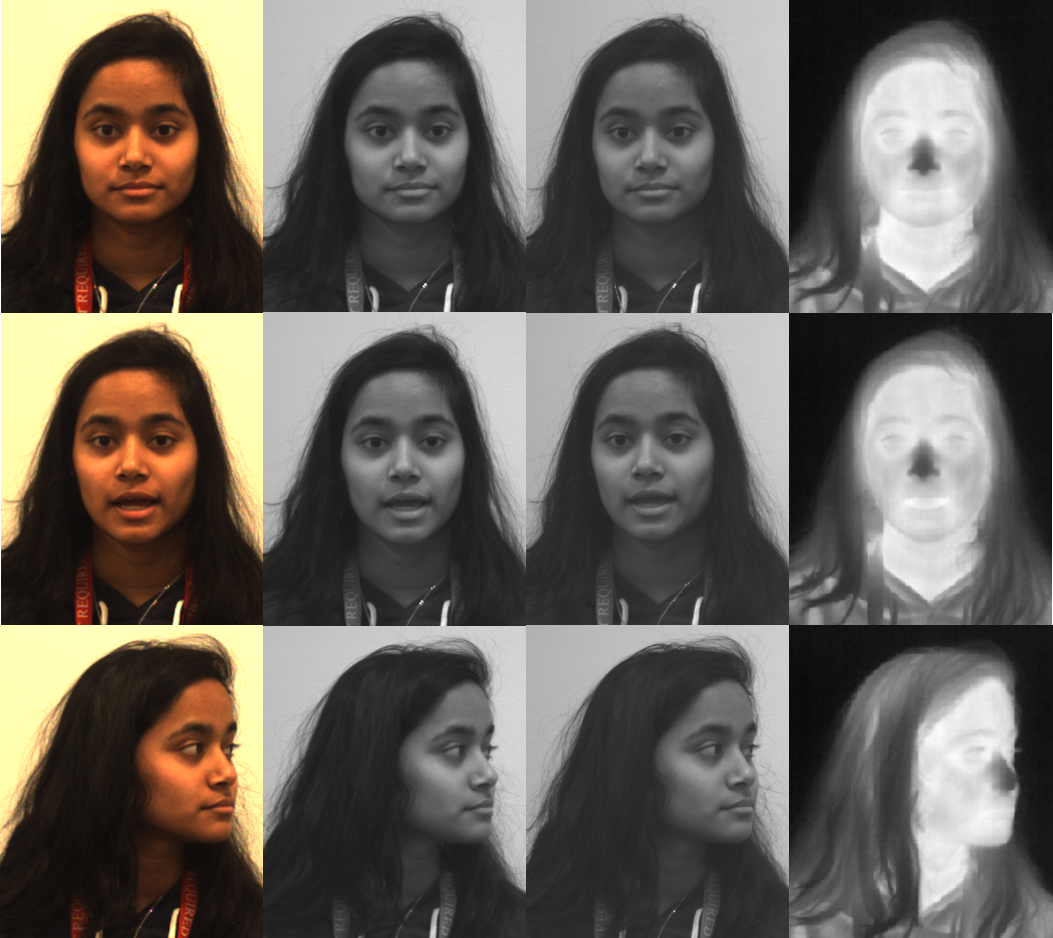}
   \vskip-10pt \caption{A set of images from the RGB (left), stereo monochrome (middle), and LWIR (right) cameras from the baseline (top), expression (middle), and off-pose (bottom) sequences.}
    \label{fig:example_images}
\end{figure}

The use of thermal imaging has grown steadily over the past several decades, aided by improvements in sensor technology as well as reductions in cost.  Thermal infrared sensors capture heat emissions, such as those radiated by the human body, in the \SI{3}{\micro\metre}-\SI{5}{\micro\metre} medium wave infrared (MWIR) band and \SI{7}{\micro\metre}-\SI{14}{\micro\metre} longwave infrared (LWIR) band.  Thermal imaging of faces have applications in the military and law enforcement for face recognition in low-light and nighttime environments \cite{Hu2017}\cite{Klare2013}\cite{Tsiamyrtzis2007}\cite{Xing_BTAS2018} and healthcare \cite{Gault2013}\cite{Pavlidis2001}\cite{Wang2020}, which require robust recognition models in challenging unconstrained operational conditions. However, the majority of MWIR and LWIR face datasets available at the time of this paper's writing consist of lower resolution images from older thermal sensors.  

While good rank-1 face recognition rates (around 90\%) have been reported using 64$\times$64 cropped face images captured by these older thermal cameras \cite{Mostafa2013}, there is still a large gap in meeting the aforementioned requirements for military, law enforcement, and healthcare applications. To help address these requirements, face datasets containing high resolution thermal imagery under various conditions, such as variable pose, expression, occlusion, and resolutions are needed. Furthermore, it is oftentimes desirable to synchronize and co-register the data being collected across multiple sensors to support algorithm development of fusion, domain adaptation, and cross-modal image synthesis approaches.

To this end, we present the Army Research Laboratory Visible-Thermal Face (ARL-VTF) dataset. This dataset is, to the best of our knowledge, the largest thermal face dataset publicly available for scientific research to date. The main contributions of the ARL-VTF dataset are:

\begin{compactitem}
    \item A multi-modal, time synchronized acquisition of 395 subjects and over 500,000 face images captured using multiple visible cameras for stereo 3D vision and one LWIR sensor (sample images shown in Figure \ref{fig:example_images}).
    \item Three image sequences capturing baseline, expression, and pose conditions for each subject. A fourth condition, eye glasses, is captured if a subject wears glasses.
    \item Annotations for head pose, eyewear, face bounding box, and 6 face landmarks locations. 
    \item Standardized protocols for model training and evaluation.
\end{compactitem}

Results and analysis on the tasks of thermal face landmark detection and thermal-to-visible face verification using state-of-the-art deep learning models are presented as a benchmark.


\section{Literature Review}
\begin{table*}[htb]
\centering
\caption{Summary statistics of datasets containing MWIR or LWIR face data ordered (approximately) from least to most recent. Whether controlled or uncontrolled, the presence of the following variable conditions is noted: (P)ose, (I)llumination, (E)xpression, (T)ime-lapse, (G)lasses, and (O)cclusion.  Image resolution is written as (w$\times$h).}
\label{tab:datasets}
\begin{tabular}{LLLLLL}
\toprule
\multicolumn{1}{l}{Dataset} &
\multicolumn{1}{c}{Modalities} &
\multicolumn{1}{c}{Subjects} &
\multicolumn{1}{c}{Variability} &
\multicolumn{1}{c}{IR Resolution} &
\multicolumn{1}{c}{Range (m)} \\
\midrule

\nm{UND \cite{Chen2003}} & \nm{LWIR, RGB} & 241 & \nm{I,E,T} & 320\times240 & \nm{Unspecified} \\
\nm{IRIS \cite{IRIS}} & \nm{LWIR, RGB} & 30 & \nm{P,I,E} & 320\times240 & \nm{Unspecified} \\
\nm{IRIS-M3 \cite{Chang2006}}  & \nm{LWIR, RGB} & 82 & \nm{I} & 320\times240 & 1.2 \\
\nm{Terravic \cite{Terravic}} & \nm{LWIR} & 20 & \nm{P,G} & 320\times240 & \nm{Unspecified} \\
\nm{UH \cite{Buddharaju2007}} & \nm{MWIR} & 138 & \nm{P,E} & 640\times512 & \nm{Unspecified} \\
\nm{NVIE \cite{Wang2010}} & \nm{LWIR, Mono} & 215 & \nm{I,E,G} & 320\times240 & 0.75 \\
\nm{KTFE \cite{Nguyen2013}} & \nm{LWIR, RGB} & 26 & \nm{E,G} & 320\times240 & 0.85 \\
\nm{Carl \cite{Espinosa-Duro2013}} & \nm{N/LWIR, RGB} & 41 & \nm{I,E,T} & 160\times120 \nm{ (LW)} & 1.35 \\
\nm{ULFMT \cite{Ghiass2018}} & \nm{MWIR, RGB} & 238 & \nm{P,E,T,G} & 640\times512 & 1.0 \\
\nm{ARL-MMFD \cite{Hu2016a}\cite{Zhang2019}} & \nm{P-L/LWIR, RGB} & 111 & \nm{E} & 640\times480 \nm{ (LW)} & 2.5, 5.0, 7.5 \\
\nm{Eurocom \cite{Mallat2018}} & \nm{LWIR, RGB} & 50 & \nm{P,I,E,G,O} & 160\times120 & 1.5 \\
\nm{RWTH \cite{Kopaczka2019}} & \nm{LWIR} & 94 & \nm{P,E} & 1024\times768 & 0.9 \\
\nm{Tufts \cite{Panetta2020}} & \nm{N/LWIR, RGB} & 100 & \nm{P,E} & 336\times256 & 1.5 \\
\nm{ARL-VTF} & \nm{LWIR, RGB, Mono} & 395 & \nm{P,E,G} & 640\times512 & 2.1 \\

\bottomrule
\end{tabular}
\end{table*}

In this section we provide a thorough comparison of several publicly released MWIR or LWIR face datasets and briefly highlight some notable characteristics of each. Table \ref{tab:datasets} presents a high-level comparison of the key statistics of different datasets, including the ARL Visible-Thermal Face Dataset (ARL-VTF) presented in this paper.

Collected primarily in 2002 with visible and LWIR cameras, the University of Notre Dame (UND) \cite{Chen2003} dataset remains as one of the largest datasets in terms of unique identities (with 241 subjects), but has only four images per subject, and used what is now considered a very low resolution and low sensitivity uncooled microbolometer.

The IRIS \cite{IRIS} dataset has simultaneous recordings of 30 subjects in variable poses and expressions in both LWIR and visible, however no annotations included besides the subject id. The IRIS-M3 \cite{Chang2006} dataset, however, contains 88 subjects simultaneously captured under a variety of indoor and outdoor lighting conditions with not only LWIR and visible cameras but also a multi-spectral imaging module.

Two different datasets have both been referred to as the University of Houston (UH) dataset. The more recent version \cite{Buddharaju2007} contains 7,590 MWIR images from 138 subjects. A slightly older version \cite{Kakadiaris2005a} contains 88 subjects and simultaneous acquisition of visible, thermal, and range data for 3d model generation. The thermal IR camera is not specified though presumably it is the same as \cite{Buddharaju2007}.

The Natural Visible and Infrared Expression Database (NVIE) \cite{Wang2010} captures subjects displaying a wide range of emotions. Sequences of unposed expressions were elicited by having subjects, some of whom wore glasses, observe video clips. Sequences of posed expressions were also captured with all subjects with and without glasses. The LWIR and grayscale visible image streams were simultaneously recorded and manually time-synchronized. While 238 subjects participated in the collection, \cite{Wang2010} notes that there is only data for 105-112 subjects in the majority of scenarios.

Similar to \cite{Wang2010}, the KTFE dataset \cite{Nguyen2013} elicited natural displays of emotion from 26 subjects through the use of video clips. Instrumental music was used between sequences to promote a neutral emotional state. Subjects were allowed to wear glasses during the collection. The data was captured simultaneously with an InfRec R300 camera.

The Carl dataset \cite{Espinosa-Duro2013} contains time-lapse data of 41 subjects captured in four separate sessions spaced two days apart in which subjects were allowed uncontrolled natural variations in their expressions. The data was simultaneously recorded using a combined visible/LWIR camera and a separate NIR camera.

The Universit{\'{e}} Laval Face Motion and Time Lapse (ULFMT) database \cite{Ghiass2018} contains 238 subjects recorded in multiple sequences under variable conditions, including significant time-lapse on the order of two to four years. Although the data was collected from the near, short, medium and long wave infrared bands, only the MWIR data has been released to date.

The ARL Multi-Modal Face Database (MMFD) dataset is composed of two separate collections, first presented in \cite{Hu2016a} and then extended in \cite{Zhang2019}, both with simultaneously acquired visible, LWIR, and Polarimetric LWIR data. It has a combined total of 111 subjects. Unique to this dataset is the variable distances at which subjects are captured.

The Eurocom dataset\cite{Mallat2018}, with 50 subjects captured using a combined visible/LWIR camera, notably contains a wide variety of acquisition scenarios, including sequences during which the eye and mouth regions are occluded by the subject's hand.

The RWTH-Aachen \cite{Kopaczka2019} dataset contains high resolution LWIR images of 94 subjects. Each subject is captured with variable expressions and head poses (both pitch and yaw), in controlled and uncontrolled sequences. The dataset is well annotated for emotions, discrete facial actions, and face landmarks. It cannot be used on its own for thermal-to-visible face recognition due to an absence of visible data, however it can still be employed to develop thermal landmark detection algorithms.

The Tufts Face Database \cite{Panetta2020} is a multi-modal dataset with several image acquisition devices and scenarios. The scenarios involve the simultaneous capture of visible and LWIR frontal images as well as visible, NIR, LWIR images acquired with a mobile, multi-camera sensor platform being rotated in front of the subject in an arc. In both scenarios, subjects were asked to pose with a variety of expressions and also sunglasses. Also included in the dataset are images from a 3D light-field camera, 3D point cloud reconstructed facial images, and computer-generated face sketches. The dataset contains 100 subjects.

Compared to the ARL-VTF dataset with 395 subjects, the next largest high-resolution thermal face dataset, ULFMT, contains 238 subjects and features MWIR and RGB video recordings under a comprehensive set of variable conditions but lacks synchronized data. For the RWTH dataset, although it utilized a higher resolution thermal camera and provides annotations for variable expressions, it contains no visible imagery counterpart. In contrast, ARL-VTF's synchronized acquisition and stereo arrangement supports algorithm development for 3D model learning \cite{deng2018uv}, multi-modal fusion \cite{Kakadiaris2005a}, domain adaptation \cite{reale2016seeing}, and  cross-domain image synthesis \cite{he2020coupled}. Three such synthesis approaches \cite{di2019polarimetric}\cite{isola2017image}\cite{Zhang2017} for thermal-to-visible face verification are showcased in Section \ref{section:verification}.

In summary, the ARL-VTF dataset is the only dataset which has all of the following characteristics: a) time-synchronized visible and thermal imagery,  b) data collected using a current commercially available uncooled LWIR camera, c) variable expression, pose, and eyewear, d) facial landmark annotations, and e) the largest number of subjects and images to-date.

\section{Database Collection}

\begin{figure}[tb]
    \centering
    \includegraphics[width=.7\linewidth]{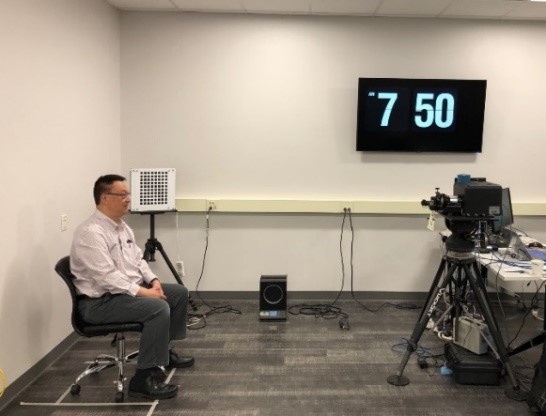}
   \vskip-10pt \caption{The collection area showing the sensor array as it collects the baseline (frontal) image sequence.}
    \label{fig:collection}
\end{figure}

The data collection occurred over the course of 9 days in November 2019.  The released dataset contains 395 subjects, each of whom completed an Institutional Review Board (IRB) approved consent form prior to image acquisition. The subjects were seated in front of a thermally neutral background 2.1 meters from the sensor array with their heads at approximately the same height as the sensors. Illumination was provided by the standard fixed overhead room lighting.  The collection area setup is pictured in Figure \ref{fig:collection}.

Subjects' faces were recorded for approximately 10 seconds under each of the following conditions:
\begin{compactenum}
  \item A \textit{baseline} sequence of frontal images with the subject maintaining a neutral expression. If subjects were wearing glasses, they were asked to remove them.
  \item An \textit{expression} sequence of frontal images of the subject counting out loud incrementally starting from one.
  \item A \textit{pose} sequence of images where subjects were asked to slowly turn their heads from left to right.  However, a small number of subjects rotated their entire bodies from left to right using the swiveling chair.
  \item If subjects naturally wear \textit{glasses} (removed for sequences 1-3), they were asked to put them back on for an additional sequence of \textit{baseline} images.
\end{compactenum}

\begin{table*}[h]
\centering
\caption{Visible and LWIR camera information. The \{$\cdot$\} enumeration corresponds to the camera labeling in Figure \ref{fig:sensors}. The Mean (M) and Standard Deviation (SD) of inter-pupil distances (IPDs) are calculated using the baseline image sequence.}
\label{tab:sensors}
\begin{tabular}{LLLLL}
\toprule
\multicolumn{1}{l}{Camera} &
\multicolumn{1}{l}{Modality} &
\multicolumn{1}{c}{\nm{Resolution (w$\times$h)}} &
\multicolumn{2}{c}{IPD} \\

\cmidrule(lr){4-5} &

\multicolumn{1}{c}{} &
\multicolumn{1}{c}{} &
\multicolumn{1}{c}{M} &
\multicolumn{1}{c}{SD} \\

\midrule

\nm{FLIR Grasshopper3 \textbf{\{1, 4\}}} & \nm{Mono visible} & 2048\times2048 & 89.3 & 6.6 \\
\nm{FLIR Boson \textbf{\{2\}}} & \nm{LWIR } \SI{7.5}-\SI{13.5}{\micro\metre} & 640\times512 & 45.2 & 3.3 \\
\nm{Basler Scout \textbf{\{3\}}} & \nm{RGB color} & 658\times492 & 66.7 & 5.0 \\
\bottomrule
\end{tabular}
\end{table*}

\begin{figure}[t]
    \centering
    \includegraphics[width=0.65\linewidth]{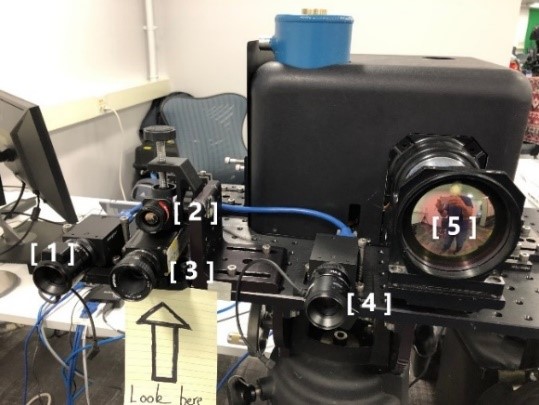}
   \vskip-7pt \caption{Sensor array with two FLIR Grasshopper3 cameras \textbf{\{1, 4\}}, the FLIR Boson LWIR sensor \textbf{\{2\}}, and the Basler Scout camera \textbf{\{3\}}. Polarimetric LWIR sensor \textbf{\{5\}} data not included.}
    \label{fig:sensors}
\end{figure}

\noindent {\bf{Sensors:}} This dataset was collected with an array of three visible cameras and one LWIR thermal sensor. The visible imagery was recorded using two monochrome FLIR Grasshopper3 CMOS cameras and one RGB Basler Scout CCD camera. The LWIR data is captured by a FLIR Boson uncooled VOx microbolometer with a spectral band of \SI{7.5}{\micro\metre} to \SI{13.5}{\micro\metre} and thermal sensitivity of $<$50 mk. Table \ref{tab:sensors} lists the camera specifications. The sensors were mounted onto a single optical plate as shown in Figure \ref{fig:sensors}. Data from a fifth sensor (a LWIR polarimeter) is omitted from this dataset as it was not time-synchronized with the other cameras.

\begin{figure}[t]
    \centering
      \begin{subfigure}{0.22\textwidth}
        \includegraphics[width=\textwidth,height=100pt]{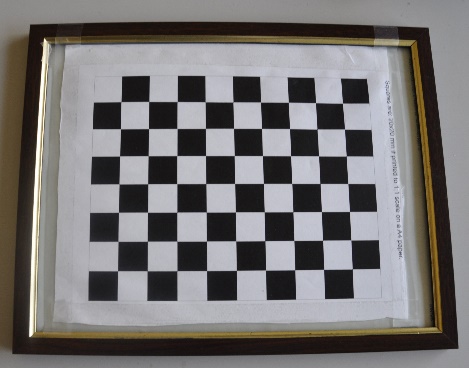}
          \caption{Visible Pattern}
          \label{fig:vis_cboard}
      \end{subfigure}
      \hfill
      \begin{subfigure}{0.22\textwidth}
        \includegraphics[width=\textwidth,height=100pt]{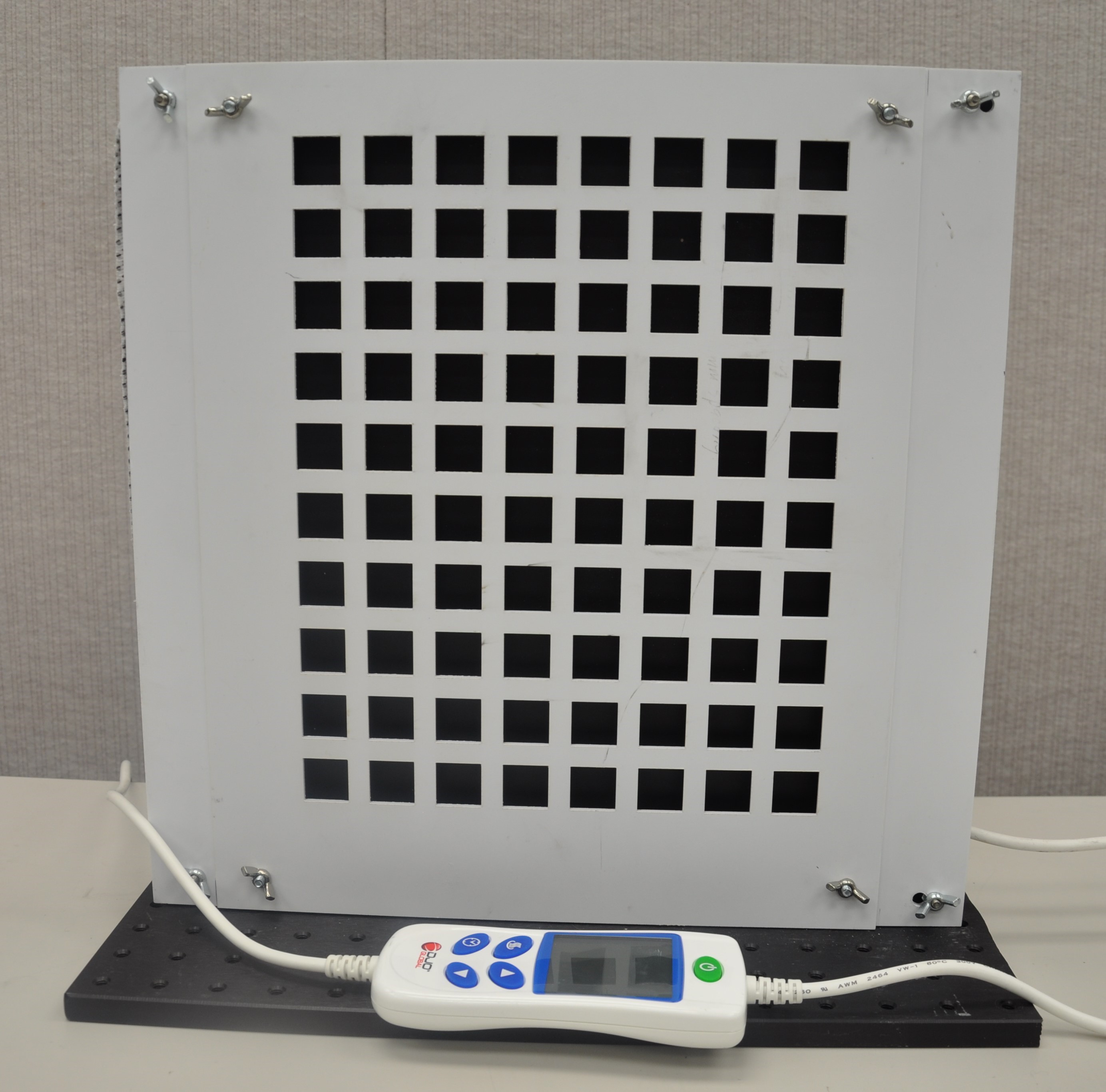}
          \caption{Thermal Pattern}
          \label{fig:ir_cboard}
      \end{subfigure}
\vskip-10pt\caption{
\label{fig:cboards}%
Calibration patterns for the visible and thermal sensors.}
\end{figure}

\noindent {\bf{Sensor Calibration and Synchronization:}}
Sensor calibrations were conducted each day of the data collection to enable post-processing for 2D image registration and 3D geometric calibrations of the multiple visible and infrared sensors. An $8\times 10$ checkerboard pattern with 20mm squares is mounted in front of a black body source which provides contrast for both visible and thermal images. For the thermal camera, a custom designed thermal/visible pattern using 20mm square holes with 10mm spacing was used. The visible and thermal sensor checkerboard calibration patterns are presented in Figure \ref{fig:cboards}. In order to facilitate the development of 3d-based algorithms, the intrinsic and extrinsic camera parameters are provided with this dataset.

Using custom software to interface with each camera vendor's respective SDK software, the images were captured in a time-synchronized fashion via multithreaded software triggers at 15 frames per second due to bandwith limitations regarding data transfer.

\subsection{Dataset Details and Usage}
In total, the dataset contains 395 subjects and 549,712 images. To provide a sense of face resolution, the average inter-pupil distances (IPDs) of frontal baseline images are tabulated in Table 3. IPDs are calculated as the pixel distance between the left and right eye centers. To facilitate reproducibility and evaluation, the dataset is divided into subject-disjoint development (training and validation) and test sets with 295 subjects in the development set and the remaining 100 subjects in the test set. The subjects within the development set are sub-divided into training and validation sets using a 5-fold cross-validation scheme for hyper-parameter tuning and model selection.  Of the 395 total subjects, 60 subjects were recorded both with and without glasses. These subjects have been evenly divided between the development and test sets, and proportionally divided between the training and validation sets (24 for training and 6 for validation).


\subsubsection{Thermal-to-Visible Face Verification Protocols}
\label{section:protocols}
We use the following grammar to describe the type of images in each gallery and probe set. In order to facilitate detailed analysis, the temporally-disjoint sets of gallery and probe images are defined in terms of a sequence category and an eyewear category. Gallery and Probe protocols are designated ``\textbf{G}" and ``\textbf{P}" respectively. ``\textbf{V}" and ``\textbf{T}" refer to the visible and thermal spectrum data. The sequence categories ``\textbf{B}", ``\textbf{E}", and ``\textbf{P}" signify the baseline, expression, and pose sequences, respectively. The ``$*$" symbol represents any or all sequence categories. For the purposes of the evaluation protocol, \textbf{B} also includes the glasses image sequence. There are three eyewear categories which describe if a subject possesses glasses and if the glasses are being worn in the image.  Images of subjects who do not possess glasses use the tag \textbf{0}, whereas subjects who have their glasses removed or worn are notated \textbf{-} and \textbf{+}, respectively. The eyewear category is omitted when no filtering has been done on the basis of eyewear. In extended Backus–Naur form, the rules for producing descriptive protocol labels are:

\setlength{\belowdisplayskip}{0pt} \setlength{\belowdisplayshortskip}{0pt}
\setlength{\abovedisplayskip}{0pt} \setlength{\abovedisplayshortskip}{0pt}

\begin{bnf*}
\bnfprod{set} {``\textbf{G}" \bnfor ``\textbf{P}"} ; \\
\bnfprod{modality} {``\textbf{V}" \bnfor ``\textbf{T}"} ; \\
\bnfprod{sequence} {``\textbf{B}" \bnfor ``\textbf{E}" \bnfor ``\textbf{P}" \bnfor ``*" ; } \\
\bnfprod{eyewear} {``\textbf{0}" \bnfor ``\textbf{-}" \bnfor ``\textbf{+}" ; } \\
\bnfprod{protocol} {\bnfpn{set}, ``\_", \bnfpn{modality}, } \\
\bnfmore{\bnfpn{sequence}, [\bnfpn{eyewear}+] ;}
\end{bnf*}

Specific protocols have been developed for the evaluation of thermal-to-visible face verification algorithms. As the collection process yielded a different number of images for each subject, the test data has been selectively sampled to provide an equal number of images per subject and sequence.  Additionally, specific images have been further designated as either probe or gallery images in order to standardize evaluation.  Gallery images are composed solely of baseline images from the visible cameras. Probes are thermal images from all three sequences. Two distinct galleries are specified: 1) \textbf{G\_VB0-} in which no subjects are wearing glasses, and 2) \textbf{G\_VB0+} wherein glasses are worn by the subjects who have them. 

The gallery and probe sets were constructed as follows. Seven evenly-spaced timestamps were selected from each subject's baseline sequence, starting from the first timestamp and ending with the last.  The images from each of the three visible cameras corresponding to the first and last timestamp in the sequence are placed into \textbf{G\_VB0-}. The images from the LWIR camera corresponding to the remaining five timestamps are designated as probes (\textbf{P\_TB0-}). If a glasses sequence was recorded for that subject, then this process is repeated for the images in that sequence, with the resulting images becoming associated with the \textbf{G\_VB0+} and \textbf{P\_TB+} protocols. Next, 25 timestamps for the expression sequence are selected, spaced evenly to cover the span of the sequence. The images corresponding to those timestamps from all four cameras are added to the subject's set of probe images (\textbf{P\_TE0-}).  The same is done for the pose sequence (\textbf{P\_TP0-}).

In summary, each subject has 6 gallery images (2 timestamps $\times$ 3 visible cameras) and 5 baseline probe images (5 timestamps $\times$ 1 thermal camera) without any eyewear. The subjects with glasses have an additional set of gallery and baseline probe images where the glasses are worn. This protocol can easily be extended to visible-to-visible or visible-to-thermal face verification by including the remaining images from the other cameras.

\begin{table}[htb]
\centering
\caption{The number of probe and gallery images in the thermal-to-visible face verification protocol per subject in the test set. Mono 1 and 2 refer to the Grasshopper3 cameras. *only pertains to subjects who brought glasses to the collection.}
\label{tab:protocols}
\begin{tabular}{LLLLL}
\toprule
\multicolumn{1}{l}{Sequence} &
\multicolumn{1}{c}{Probes} &
\multicolumn{3}{c}{Gallery} \\

\cmidrule(lr){2-2}
\cmidrule(lr){3-5} &

\multicolumn{1}{c}{LWIR} &
\multicolumn{1}{c}{Mono 1} &
\multicolumn{1}{c}{Mono 2} &
\multicolumn{1}{c}{RGB} \\
\midrule
\midrule

\nm{baseline} & 5 & 2 & 2 & 2 \\
\nm{expression} & 25 & 0 & 0 & 0 \\
\nm{pose} & 25 & 0 & 0 & 0 \\
\nm{glasses*} & 5 & 2 & 2 & 2 \\
\bottomrule
\end{tabular}
\end{table}

However, it should be noted that the development set has not been similarly balanced.  All available images of a subject are by default included in the development set.  Sub-sampling the development data is left to the user's discretion.

\noindent {\bf{Annotations:}}
Face bounding box and face landmark coordinates were generated using a commercial off-the-shelf face and landmark detector (Neurotechnology Verilook SDK) applied independently to the two high-resolution FLIR Grasshopper3 images assisted by manual supervision and correction of annotations.  Face landmarks are in a 6-point annotation scheme corresponding to the left eye center, right eye center, base of nose, left mouth corner, right mouth corner, and center of mouth.  The stereo arrangement of the Grasshopper3 cameras enabled the annotated points to be projected into the coordinate spaces of the LWIR and Scout RGB cameras using 3D geometry.  

\begin{figure}[ht]
    \centering
    \includegraphics[width=.65\linewidth]{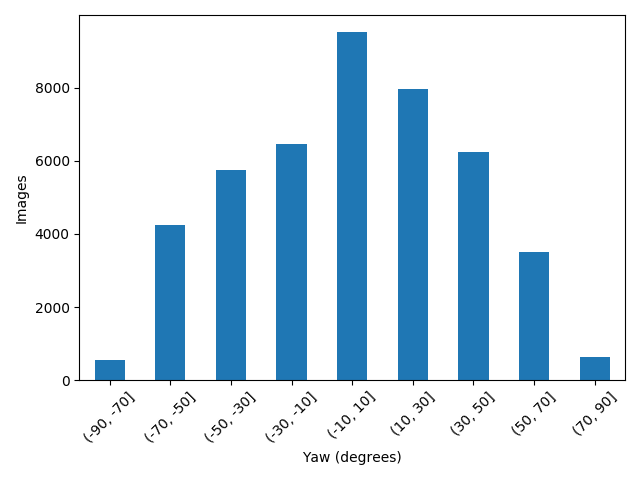}
   \vskip-10pt \caption{Distribution of head poses in terms of estimated yaw angles from the pose image sequence.}
    \label{fig:yaw_dist}
\end{figure}

The stereo setup also allowed for the automatic estimation of head pose achieved using OpenCV's \cite{opencv_library} implementation of the Perspective-n-Point with RANSAC algorithm.  Figure \ref{fig:yaw_dist} displays the distribution of estimated yaw angles captured during the pose sequence across all subjects. There is some slight asymmetry in the distribution about 0$\degree$, partially due to the fact that subjects oftentimes did not complete the full 180$\degree$ head rotation. Metadata for each image includes the subject ID, camera, timestamp, image sequence, detected face bounding box, detected 6-point face landmarks, and estimated yaw angle.\\
\noindent {\bf{Requesting the Database:}}
Requests for the database can be made by contacting Matthew Thielke (matthew.d.thielke.civ@mail.mil). Requestors will be asked to sign a database release agreement and each request will be vetted for valid scientific research.

\section{Performance Benchmarks}
Benchmark results for landmark detection and thermal-to-visible face verification are provided in this section. 

\subsection{Face Landmark Detection}

\begin{table*}[h]
\centering
\caption{Landmark detection performance statistics in terms of the NRMSE.}
\label{tab:nrmse}
\begin{tabular}{LLLLLLLL}
\toprule
\multicolumn{1}{l}{Sequence} &
\multicolumn{1}{c}{Mean} &
\multicolumn{1}{c}{Std} &
\multicolumn{1}{c}{Median} &
\multicolumn{1}{c}{MAD} &
\multicolumn{1}{c}{Max Error} &
\multicolumn{1}{c}{AUC$_{0.08}$} &
\multicolumn{1}{c}{Failure Rate$_{0.08}$} \\
\midrule

\nm{baseline} & 0.032581 & 0.015483 & 0.0283 & 0.0119 & 0.0857 & 0.5798 & 0.0080 \\
\nm{expression} & 0.032445 & 0.015679 & 0.0276 & 0.0122 & 0.1109 & 0.5946 & 0.0076 \\
\nm{glasses} & 0.036763 & 0.014687 & 0.0350 & 0.0114 & 0.0737 & 0.4649 & 0.0000 \\
\nm{pose} & 0.101184 & 0.056227 & 0.0949 & 0.0472 & 0.4431 & 0.1692 & 0.5868 \\
\bottomrule
\end{tabular}
\end{table*}

The Deep Alignment Network (DAN) \cite{Kowalski2017e} is a multi-stage convolutional neural network (CNN) designed to iteratively update the predicted landmark locations given an initial shape estimate.  It has shown promising results for face landmark detection on both visible \cite{Zafeiriou2017c} and thermal \cite{Poster2019}\cite{Kopaczka2019} imagery. The model was trained with thermal face images from all of the recording sequences. The detected face bounding boxes are used to crop the images. The output of the model is the predicted face shape $\mathbf{\hat{y}} \in \mathbb{R}^{L \times 2}$, where $L$ is the number of face landmark locations.

For these benchmarks, we set $L=5$ and detect the left and right eye centers, the base of the nose, and the left and right mouth corners. Landmark detection performance is evaluated using the Normalized Root Mean Square Error (NRMSE),
\begin{equation}
    E(\mathbf{\hat{y}}, \mathbf{y}) = \frac{1}{N}\sum_{i=1}^{N} \frac{\frac{1}{L}\sum_{j=1}^{L} \left \|\mathbf{\hat{y}}_{i,j} - \mathbf{y}_{i,j}\right \|_{2}}{\left \|tl_i - br_i\right \|_2},
\end{equation}

where $N$ is the number of samples in the test set and $\mathbf{\hat{y}}$ and $\mathbf{y}$ are the predicted and ground-truth landmark coordinates, respectively. The error is normalized by the Euclidean distance between the top left point, $tl$, and bottom right point, $br$, of the ground-truth shape's rectangular bounds. The face diagonal is used to normalize the error, rather than the IPD, as it is more stable in off-pose conditions \cite{wu2019facial}. As per \cite{Zafeiriou2017c}, in addition to the mean and standard deviation (Std), the median, Median Absolute Deviation (MAD), and maximum NRMSE statistics are tabulated in Table \ref{tab:nrmse}. We set a threshold of 0.08 NRMSE for the Failure Rate and Area Under the Curve (AUC) of the Cumulative Error Distribution (CED).

\begin{figure}[h]
    \centering
    \includegraphics[width=.7\linewidth]{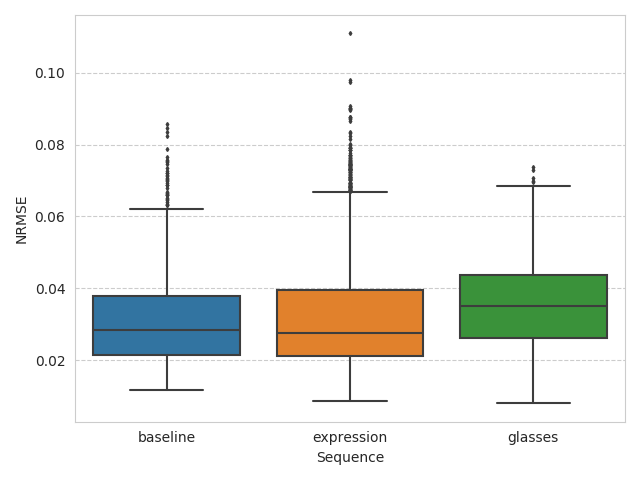}
   \vskip-10pt \caption{NRMSE of baseline, expression, and glasses sequences.}
    \label{fig:nrmse_box}
\end{figure}

\begin{figure}[h]
    \centering
    \includegraphics[width=.7\linewidth]{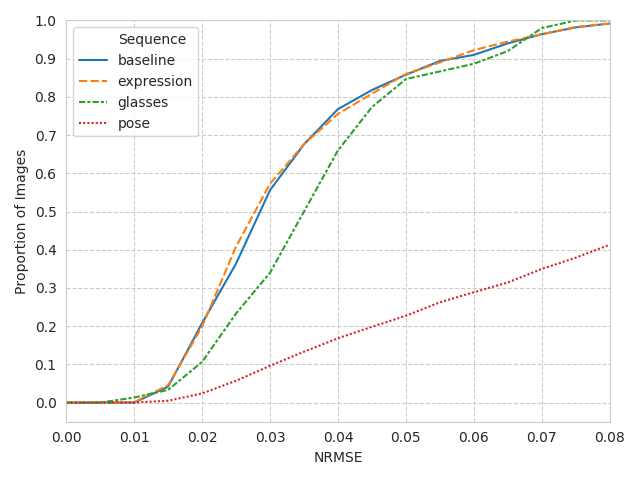}
   \vskip-10pt \caption{CED for the baseline, expression, glasses, and pose sequences.}
    \label{fig:ced}
\end{figure}

\begin{figure}[h]
    \centering
    \includegraphics[width=.65\linewidth]{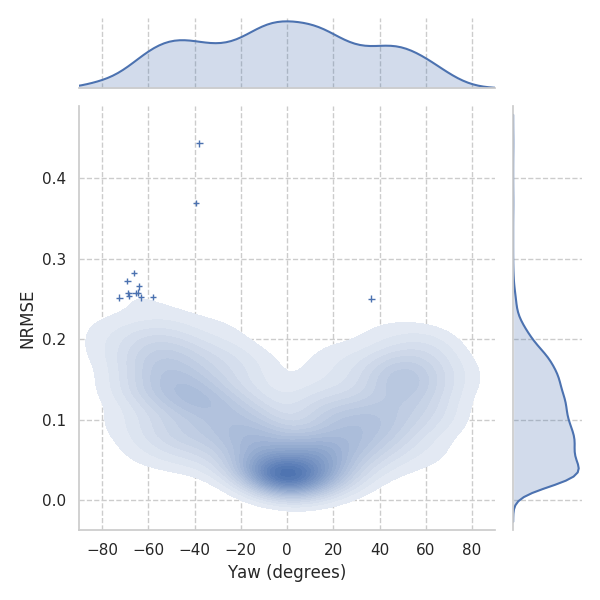}
   \vskip-10pt \caption{Bivariate distribution generated using Gaussian kernel density estimator of NRMSE across head yaw for the pose sequence. `$+$' indicates outliers with NRMSE $\geq 0.24$.}
    \label{fig:yaw_nrmse}
\end{figure}

As seen from Figures \ref{fig:nrmse_box} and \ref{fig:ced}, the DAN achieves good performance on all frontal images, including images with expressions or glasses. The model fails on the head pose sequence, where performance significantly degrades with yaw angles beyond $\pm$20$\degree$, as illustrated in Figure \ref{fig:yaw_nrmse}. Interestingly, while images with glasses have a slightly higher NRMSE on average compared to the other frontal images, they also have tighter performance bounds and a 0\% Failure Rate, as shown in Figure \ref{fig:nrmse_box}. This may be due to the distinct visual cues granted by glasses (which absorb heat emissions and appear black in thermal images), or simply by virtue of a small sample size of subjects with glasses.

\subsection{Thermal-to-Visible Face Verification}\label{section:verification}
One domain-invariant feature learning approach and three thermal-to-visible synthesis approaches are benchmarked against the ARL-VTF dataset. The verification performance is measured by the Receiver Operating Characteristic (ROC) curve and the Area Under the Curve (AUC) metrics, as well as the True Accept Rate (TAR) at False Accept Rates (FAR) equaling 1\% and 5\%.

\begin{figure}[]
    \centering
    \includegraphics[width=\linewidth]{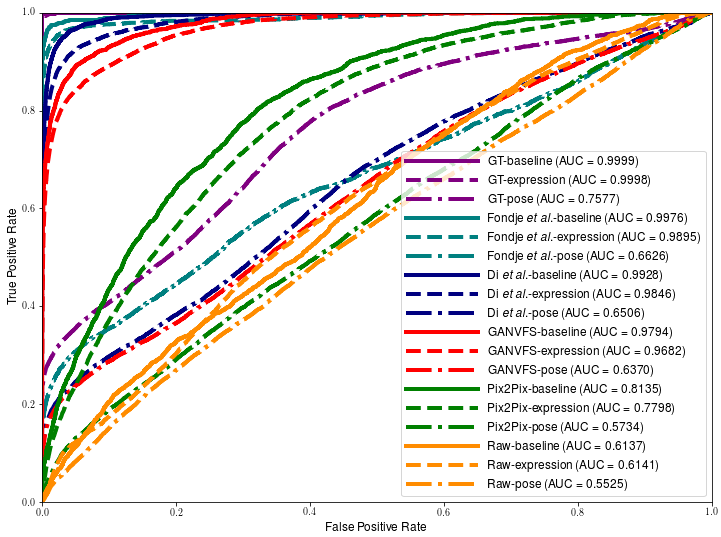}
   \vskip-10pt \caption{The ROC curves corresponding to the different methods for gallery \textbf{G\_VB0-} and protocols \textbf{P\_T*0-}.}
    \label{fig:roc}
\end{figure}

The first method matches thermal and visible face images by learning a domain adaptive feature extractor as proposed in \cite{fondje2020cross}. This framework exhibits four main parts: (1) a truncated version of VGG16 and Resnet to extract common features, (2) a “Residual Spectral Transform” subnetwork that learns a mapping between the visible and thermal features, (3) a cross-domain identification loss to optimize task-level discrimination, and (4) a domain invariance loss which ensures domain unpredictability. The extracted probe and gallery image features are compared using the cosine similarity measure. The results reported in Figure~\ref{fig:roc} and Table~\ref{tab:comparison_table} corresponding to this baseline were yielded by the VGG16 version of the framework. The images are preprocessed similarly to \cite{Riggan2016}\cite{Hu2016a} (with bandpass filtering omitted) by first aligning images to a 5-point canonical coordinate scheme via similarity transformation and then loosely cropping the aligned face images to $360\times280$ pixels in order to provide enhanced contextual information.

The remaining three methods employ Generative Adversarial Networks (GANs) to learn a mapping from thermal face images to visible face images. Once the visible image is synthesized from the input probe thermal image, a pre-trained VGG-Face model \cite{parkhi2015deep} is used to extract deep features (i.e. output from relu5\_3 layer ) from the synthesized visible probe image as well as the visible gallery image to perform thermal to visible face verification. The cosine similarity between the two feature vectors is calculated to produce the verification score. The inputs into these synthesis models are $128 \times 128$ face images cropped according to the annotated bounding boxes. Images from all four sequences are used to train the models. The following GAN-based methods are used for evaluation:

\begin{compactitem}
    \item Pix2Pix \cite{isola2017image}: Conditioned on thermal images, Pix2Pix model synthesizes visible images using a U-net based architecture \cite{isola2017image}\cite{ronneberger2015u}.
    \item GANVFS \cite{Zhang2017}: GANVFS uses identity loss and perceptual loss \cite{johnson2016perceptual} to train a synthesis network.
    \item Self-attention based CycleGAN (SAGAN) \cite{di2019polarimetric}: A self-attention module \cite{zhang2019self} is adapted with CycleGAN \cite{zhu2017unpaired} for thermal to visible synthesis.
\end{compactitem}

\begin{table*}[t]
    \caption{Verification performance comparisons among the baseline methods, state-of-the-art methods for various settings.}
    \centering
    \resizebox{2\columnwidth}{!}{%
    \begin{tabular}{LLLLLLLLLL}
    \toprule 
    \multicolumn{1}{c}{} & \multicolumn{1}{l}{} & \multicolumn{4}{c}{Gallery G\_VB0-} & \multicolumn{4}{c}{Gallery G\_VB0+} \\
    \cmidrule(lr){3-6} \cmidrule(lr){7-10}
    \multicolumn{1}{l}{Probes} & \multicolumn{1}{l}{Method} & \nm{AUC} & \nm{EER} &  \nm{FAR=1\%} & \nm{FAR=5\%} & \nm{AUC} & \nm{EER} &  \nm{FAR=1\%} & \nm{FAR=5\%} \\
    
    \midrule  {\multirow{4}{*}{P\_TB0}}
    & \nm{Raw} & 61.37 & 43.36 & 3.13  & 11.28 & 62.83 & 42.37 & 4.19 & 13.29 \\
    & \nm{Pix2Pix \cite{isola2017image}} & 71.12 & 33.80  &  6.95 & 21.28 & 75.22 & 30.42 & 8.28 & 27.63 \\
    & \nm{GANVFS} \cite{Zhang2017} & 97.94 & 8.14 & 75.00  & 88.93 & 98.58 & 6.94 & 79.09 & 91.04 \\
    & \nm{Di \etal \cite{di2019polarimetric}} & 99.28 & 3.97 & 87.95  & 96.66 & 99.49 & 3.38 & 90.52 & 97.81 \\
    & \nm{Fondje \etal \cite{fondje2020cross}} & 99.76 & 2.30 & 96.84 & 98.43 & 99.87 & 1.84 & 97.29 & 98.80 \\
    \cdashline{1-2}
    {\multirow{1}{*}{P\_VB0}}
    & \nm{GT Vis-to-Vis} & 99.99 & 0.23 & 99.79  & 99.95 & 99.99  & 0.24 & 99.86 & 100.00 \\
    \cmidrule(lr){1-10}
    
    {\multirow{4}{*}{P\_TB-}}
    & \nm{Raw} & 61.14 & 41.64 & 2.77  & 16.11 & 57.61 & 44.73 & 1.38 & 6.11 \\
    & \nm{Pix2Pix} \cite{isola2017image} & 68.77 & 38.02  &  6.69 & 20.28 & 52.11 & 48.88 & 2.22 & 4.66 \\
    & \nm{GANVFS} \cite{Zhang2017} & 99.36 & 3.77 & 84.88  & 97.66 & 87.34 & 18.66 & 7,00 & 29.66 \\
    & \nm{Di \etal \cite{di2019polarimetric}} & 99.63 & 2.66 & 91.55  & 98.88 & 89.24 & 19.49 & 16.33 & 41.22 \\
    & \nm{Fondje \etal \cite{fondje2020cross}} & 99.83 & 1.95 & 96.00 & 99.48 & 99.03 & 4.79 & 85.56 & 95.86 \\
    \cdashline{1-2}
     {\multirow{1}{*}{P\_VB-}}
    & \nm{GT Vis-to-Vis} & 100.00 & 0.00 & 100.00  & 100.00 & 99.06  & 4.33 & 89.66 & 96.22 \\
    \cmidrule(lr){1-10}

    
    \end{tabular}
    }
    \label{tab:comparison_table}
\end{table*}

\begin{figure}[]
    \centering
    \includegraphics[width=.65\linewidth]{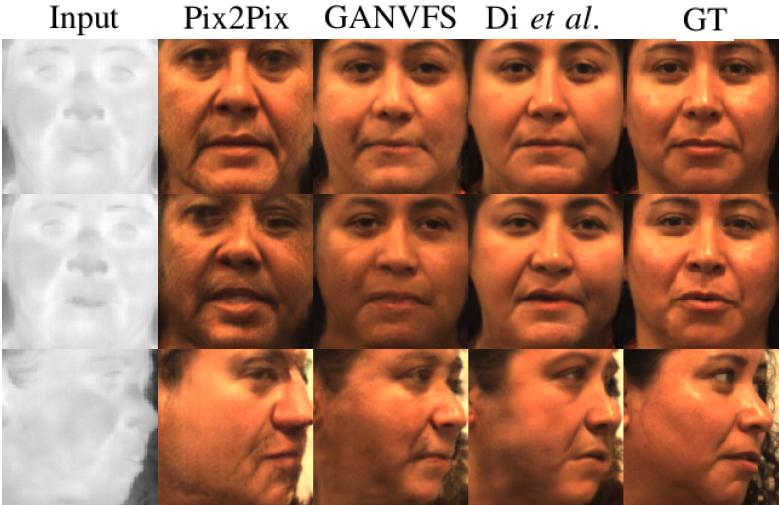}
   \vskip-10pt \caption{Sample synthesized images corresponding to different methods. First, second, and third rows correspond to baseline, expression, and profile faces.}
    \label{fig:gan_example}
\end{figure}

Additionally, two baseline methods are established to gauge the performance of the GAN-based approaches. As a naive baseline method (labelled ``Raw"), the thermal probes and visible gallery images are input directly to the VGG-Face model. In this scenario, no synthesis is performed on the thermal probes, nor is the VGG-Face model trained on the thermal data. As a ground-truth baseline method (labelled ``GT”), the thermal probe images are replaced with the corresponding ``ground-truth" visible images captured synchronously by the Basler Scout RGB camera.

The cross-modal face verification and synthesis results are shown in Figure \ref{fig:roc} and Figure \ref{fig:gan_example}, respectively.  As can be seen from Figure \ref{fig:roc}, simply extracting deep features from the raw images does not produce good verification results. This is mainly due to fact that both thermal and visible images have significantly different characteristics. The AUC corresponding to this method is only 61.37\%. Pix2Pix which is a conditional GAN-based method provides slightly better results than the simple baseline of extracting features from raw data producing AUC of 71.12\%. Both GANVFS and SAGAN methods are more advanced synthesis approaches and perform much better on this dataset, producing AUC of 97.94\% and 99.28\%, respectively. The Equal Error Rates (EER) of the Pix2Pix, GANVFS, and SAGAN models are 33.8\%, 8.14\%, and 3.97\%, respectively. The synthesis results shown in Figure \ref{fig:gan_example} are also consistent with the verification results shown in Figure \ref{fig:roc} and Table~\ref{tab:comparison_table}.

In addition to the baseline comparisons, we analyze how different variations (baseline, expression, pose, eyewear) influence the cross-spectrum matching performance of different methods. As can be seen from Figure \ref{fig:gan_example}, expression slightly degrades the performance of the baseline methods. For instance, the AUC performance of SAGAN method reduces from 99.28\% to 98.46\%. We see similar degradation for GANVFS and Pix2Pix methods on expressive face images as well. From Figures \ref{fig:roc} and \ref{fig:gan_example}, we can also see that pose affects the performance of different synthesis methods the most. The performance of the synthesis-based methods is constrained by the VGG-Face model's performance. This is evidenced by a reduction of the AUC from 99.99\% in the baseline sequence to 75.76\% for the pose sequence when using the ground-truth visible probe images as input. The EER of the Pix2Pix, GANVFS, and SAGAN models are 47.22\%, 41.66\%, and 40.24\%, respectively. This experiment clearly shows that there is much that need to be done to deal with pose, expression and occlusion variations for cross-modal synthesis and verification. More advanced methods that specifically address these issues for heterogeneous face synthesis and verification are needed. A complete set of performance metrics for all the models, probe sets, and galleries are included in supplementary material.



\section{Conclusion}
A new, large-scale face dataset of time-synchronized visible and LWIR thermal imagery is presented. In order to emulate real-world conditions, variations of expressions, head pose, and eyeglasses have been systematically captured. Furthermore, the dataset is evaluated on the tasks of thermal face landmark detection and thermal-to-visible face verification using multiple state-of-the-art algorithms. Analysis of the results indicates two challenging scenarios. First, the performance of the thermal landmark detection and thermal-to-visible face verification models were severely degraded on off-pose images. Secondly, the thermal-to-visible face verification models encountered an additional challenge when a subject was wearing glasses in one image but not the other. This effect is further exacerbated in the thermal domain due to the occlusion induced by heat absorption in the lenses.

\noindent {\bf{Acknowledgements:}}
The authors would like to acknowledge sponsorship provided by the Defense Forensics \& Biometrics Agency to conduct this research, and thank Michelle Giorgilli and Tom Cantwell for the discussions and their guidance. The authors would also like to thank Lars Ericson at IARPA and Chris Nardone, Marcia Patchan, and Stergios Papadakis at the JHU Applied Physics Laboratory for enabling ARL's participation in the 2019 IARPA ODIN data collection.

{\small
\bibliographystyle{ieee_fullname}
\bibliography{egbib}
}

\newpage

\begin{table*}[t]
    \caption{Verification performance comparisons among the baseline methods, state-of-the-art methods for various settings.}
    \centering
    \resizebox{2\columnwidth}{!}{%
    \begin{tabular}{LLLLLLLLLL}
    \toprule 
    \multicolumn{1}{c}{} & \multicolumn{1}{l}{} & \multicolumn{4}{c}{Gallery G\_VB0-} & \multicolumn{4}{c}{Gallery G\_VB0+} \\
    \cmidrule(lr){3-6} \cmidrule(lr){7-10}
    \multicolumn{1}{l}{Probes} & \multicolumn{1}{l}{Method} & \nm{AUC} & \nm{EER} &  \nm{FAR=1\%} & \nm{FAR=5\%} & \nm{AUC} & \nm{EER} &  \nm{FAR=1\%} & \nm{FAR=5\%} \\
    
    \midrule  {\multirow{4}{*}{P\_TB0}}
    & \nm{Raw} & 61.37 & 43.36 & 3.13  & 11.28 & 62.83 & 42.37 & 4.19 & 13.29 \\
    & \nm{Pix2Pix \cite{isola2017image}} & 71.12 & 33.80  &  6.95 & 21.28 & 75.22 & 30.42 & 8.28 & 27.63 \\
    & \nm{GANVFS} \cite{Zhang2017} & 97.94 & 8.14 & 75.00  & 88.93 & 98.58 & 6.94 & 79.09 & 91.04 \\
    & \nm{Di \etal \cite{di2019polarimetric}} & 99.28 & 3.97 & 87.95  & 96.66 & 99.49 & 3.38 & 90.52 & 97.81 \\
    & \nm{Fondje \etal} & 99.76 & 2.30 & 96.84 & 98.43 & 99.87 & 1.84 & 97.29 & 98.80 \\
    \cdashline{1-2}
    {\multirow{1}{*}{P\_VB0}}
    & \nm{GT Vis-to-Vis} & 99.99 & 0.23 & 99.79  & 99.95 & 99.99  & 0.24 & 99.86 & 100.00 \\
    \cmidrule(lr){1-10}
    
    {\multirow{4}{*}{P\_TB-}}
    & \nm{Raw} & 61.14 & 41.64 & 2.77  & 16.11 & 57.61 & 44.73 & 1.38 & 6.11 \\
    & \nm{Pix2Pix} \cite{isola2017image} & 68.77 & 38.02  &  6.69 & 20.28 & 52.11 & 48.88 & 2.22 & 4.66 \\
    & \nm{GANVFS} \cite{Zhang2017} & 99.36 & 3.77 & 84.88  & 97.66 & 87.34 & 18.66 & 7,00 & 29.66 \\
    & \nm{Di \etal \cite{di2019polarimetric}} & 99.63 & 2.66 & 91.55  & 98.88 & 89.24 & 19.49 & 16.33 & 41.22 \\
    & \nm{Fondje \etal} & 99.83 & 1.95 & 96.00 & 99.48 & 99.03 & 4.79 & 85.56 & 95.86 \\
    \cdashline{1-2}
     {\multirow{1}{*}{P\_VB-}}
    & \nm{GT Vis-to-Vis} & 100.00 & 0.00 & 100.00  & 100.00 & 99.06  & 4.33 & 89.66 & 96.22 \\
    \cmidrule(lr){1-10}
    
    {\multirow{4}{*}{P\_TE0}}
    & \nm{Raw} & 61.40 & 41.96 & 3.40  & 12.18 & 62.50 & 41.38 & 4.60 & 13.25 \\
    & \nm{Pix2Pix} \cite{isola2017image} & 69.10 & 35.98  &  7.01 & 16.44 & 73.97 & 31.87 & 7.93 & 19.60 \\
    & \nm{GANVFS} \cite{Zhang2017} & 96.81 & 10.51 & 70.41  & 84.00 & 97.73 & 8.90 & 74.20 & 86.80 \\
    & \nm{Di \etal \cite{di2019polarimetric}} & 98.46 & 6.44 & 81.11  & 92.49 & 98.89 & 5.60 & 84.23 & 93.94 \\
    & \nm{Fondje \etal} & 98.95 & 3.61 & 92.61 & 96.88 & 99.01 & 3.57 & 92.69 & 96.93 \\
    \cdashline{1-2}
     {\multirow{1}{*}{P\_VE0}}
    & \nm{GT Vis-to-Vis} & 99.98 & 5.38  & 99.65 & 99.92  & 99.98 & 0.45 & 99.73 & 99.96 \\
    \cmidrule(lr){1-10}
    
    {\multirow{4}{*}{P\_TE-}}
    & \nm{Raw} & 63.26 & 42.34 & 4.66  & 16.28 & 59.33 & 43.17 & 2.04 & 8.00 \\
    & \nm{Pix2Pix} \cite{isola2017image} & 68.78 & 36.24  &  7.75 & 18.06 & 51.05 & 49.11 & 2.26 & 4.95 \\
    & \nm{GANVFS} \cite{Zhang2017} & 98.66 & 5.93 & 73.17  & 92.82 & 83.68 & 22.41 & 6.77 & 22.13 \\
    & \nm{Di \etal \cite{di2019polarimetric}} & 99.30 & 3.84 & 82.55  & 97.44 & 86.12 & 21.68 & 9.88 & 31.62 \\
    & \nm{Fondje \etal} & 99.83 & 2.27 & 95.66 & 99.48 & 99.48 & 3.05 & 89.45 & 98.07 \\
    \cdashline{1-2}
    {\multirow{1}{*}{P\_VE-}}
    & \nm{GT Vis-to-Vis} & 99.99 & 0.14  & 99.97 & 99.97  & 97.96 & 6.69 & 72.16 & 90.91 \\
    \cmidrule(lr){1-10}
    
    {\multirow{4}{*}{P\_TP0}}
    & \nm{Raw} & 55.24 & 46.25 & 2.23  & 8.25 & 55.10 & 46.34 & 2.91 & 8.74 \\
    & \nm{Pix2Pix} \cite{isola2017image} & 54.86 & 47.22  &  3.13 & 9.78 & 56.50 & 46.03 & 4.01 & 10.84 \\
    & \nm{GANVFS} \cite{Zhang2017} & 63.70 & 41.66 & 16.55  & 23.73 & 65.58 & 40.19 & 17.95 & 25.68 \\
    & \nm{Di \etal \cite{di2019polarimetric}} & 65.06 & 40.24 & 17.33  & 24.56 & 67.13 & 38.67 & 18.91 & 26.46 \\
    & \nm{Fondje \etal} & 66.26 & 38.05 & 22.18 & 30.72 & 68.39 & 36.86 & 22.64 & 31.81 \\
    \cdashline{1-2}
    {\multirow{1}{*}{P\_VP0}}
    & \nm{GT Vis-to-Vis} & 75.76 & 32.30  & 28.54 & 35.52  & 77.24 & 30.92 & 29.49 & 37.27 \\
    \cmidrule(lr){1-10}
    
    {\multirow{4}{*}{P\_TP-}}
    & \nm{Raw} & 55.48 & 45.98 & 3.25  & 8.47 & 56.82 & 44.74 & 2.09 & 7.57 \\
    & \nm{Pix2Pix} \cite{isola2017image} & 54.31 & 47.04  &  2.93 & 8.44 & 50.08 & 49.67 & 0.60 & 4.33 \\
    & \nm{GANVFS} \cite{Zhang2017} & 65.79 & 40.35 & 17.84  & 25.48 & 59.51 & 44.04 & 4.29 & 15.47 \\
    & \nm{Di \etal \cite{di2019polarimetric}} & 67.27 & 39.00 & 18.16  & 26.02 & 60.10 & 43.57 & 5.77 & 15.97 \\
    & \nm{Fondje \etal} & 68.24 & 37.60 & 23.09 & 33.54 & 63.29 & 41.79 & 18.79 & 27.93 \\
    \cdashline{1-2}
    {\multirow{1}{*}{P\_VP-}}
    & \nm{GT Vis-to-Vis} & 75.59 & 33.37  & 29.37 & 36.64  & 69.62 & 37.61 & 19.36 & 28.11 \\
    \cmidrule(lr){1-10}
    
    {\multirow{4}{*}{P\_TB+}}
    & \nm{Raw} & 59.52 & 42.60 & 4.66  & 6.00 & 78.26 & 29.77 & 3.88 & 21.33 \\
    & \nm{Pix2Pix} \cite{isola2017image} & 59.68 & 41.72  &  3.33 & 3.33 & 67.08 & 36.44 & 2.68 & 11.11 \\
    & \nm{GANVFS} \cite{Zhang2017} & 87.61 & 20.16 & 20.55  & 44.66 & 96.82 & 8.66 & 46.77 & 83.00 \\
    & \nm{Di \etal \cite{di2019polarimetric}} & 91.11 & 17.43 & 22.33  & 55.66 & 97.96 & 7.21 & 60.11 & 88.70 \\
    & \nm{Fondje \etal}  & 99.28 & 5.32 & 89.21 & 94.79 & 99.97 & 0.73 & 99.47 & 100.00 \\
    \cdashline{1-2}
    {\multirow{1}{*}{P\_VB+}}
    & \nm{GT Vis-to-Vis} & 99.62 & 2.70  & 93.22 & 98.51  & 99.92 & 1.44 & 98.55 & 99.66 \\
    \bottomrule
    
    \end{tabular}
    }
    \label{tab:comparison_table}
\end{table*}
\end{document}